%% file: main.tex
\documentclass[runningheads,final]{llncs}
\usepackage{caption}
\usepackage{subcaption}
\usepackage{graphicx}
\usepackage[urlcolor=blue,hidelinks]{hyperref}
\usepackage[shortlabels]{enumitem}
\newcommand{\IIPEcosphere}[0]{IIP-Ecosphere}

\usepackage{xcolor}
\definecolor{black}{rgb}		{0.0, 0.0, 0.0}
\definecolor{white}{rgb}		{1.0, 1.0, 1.0}
\definecolor{yellow}{rgb}		{1.0, 1.0, 0.8}
\definecolor{red}{rgb}			{0.6, 0.0, 0.2}
\definecolor{blue}{rgb}		{0.0, 0.2, 0.5}
\definecolor{green}{rgb}		{0.6, 0.8, 0.8}
\definecolor{dark_green}{RGB} {0, 140, 0}
\definecolor{gold}{rgb}		{0.6, 0.4, 0.1}
\definecolor{grey}{RGB}{0,0,0}
\definecolor{Gray}{gray}{0.8}
\definecolor{MediumGray}{gray}{0.9}
\definecolor{LightGray}{gray}{0.98}
\definecolor{LightCyan}{rgb}{0.88,1,1}
\definecolor{purple}{RGB}{128,0,128}
\definecolor{slblue}{RGB}{47, 60, 105}
\definecolor{orange}{RGB}{255,165,0}
\definecolor{Gray}{gray}{0.85}

\newcommand{\comHE}[1]{\textcolor{purple}{\textbf{\large [}\colorbox{yellow}{\textbf{Holger:}}{\small #1}\textbf{\large ]}}}
\newcommand{\eg}{e.\,g.,\ }
\newcommand{\ie}{i.\,e.,\ }
\newcommand{\wrt}{w.\,r.\,t.\ }

\usepackage{dirtytalk}
\usepackage{booktabs}
\usepackage{changes}
\definechangesauthor[name={Greg}, color=blue]{GP}
\definechangesauthor[name={Holger}, color=purple]{HE}
\definechangesauthor[name={Sofiane}, color=orange]{SL}

\begin{document}
\title{Developing an AI-enabled IIoT platform - Lessons learned from  early use case validation\thanks{Supported by the German Ministry of Economics and Climate Action (BMWK) under grant numbers 01MK20006A and 01MK20006D.}}
\author{Holger Eichelberger\inst{1}
%\orcidID{0000-0002-2584-5558} 
\and Gregory Palmer\inst{2}
%\orcidID{1111-2222-3333-4444}
\and Svenja Reimer\inst{3}
%\orcidID{2222--3333-4444-5555} 
\and {Tat Trong Vu}\inst{2}
\and\\ {Hieu Do}\inst{2}
\and {Sofiane Laridi}\inst{2}
\and Alexander Weber\inst{1} 
\and Claudia Niederée\inst{2}
\and\\ Thomas Hildebrandt\inst{4}}
\authorrunning{H. Eichelberger, G. Palmer, S. Reimer, et al.}
\titlerunning{Developing an AI-enabled IIoT platform - An early use case validation}
\institute{University of Hildesheim, Software Systems Engineering, Universitätsplatz 1, 31141 Hildesheim, Germany %\email{\{eichelberger,weber\}@sse.uni-hildesheim.de}
\and
University of Hannover, L3S, Appelstraße 9a, 30617 Hannover, Germany
%\email{\{palmer, niederee\}l3s@de}
\and
University of Hannover, IFW, An der Universität 2, 30832 Garbsen, Germany\\
%\email{reimer@ifw.uni-hannover.de}
\and 
Phoenix Contact, Flachsmarktstraße 8, 32825 Blomberg, Germany
}
\maketitle              % typeset the header of the contribution
\begin{abstract}
For a broader adoption of AI in industrial production, adequate infrastructure capabilities are crucial. This includes easing the integration of AI with industrial devices, support for distributed deployment, monitoring, and consistent system configuration.

Existing IIoT platforms still lack required capabilities to flexibly integrate reusable AI services and relevant standards such as Asset Administration Shells or OPC UA in an open, ecosystem-based manner. This is exactly what our \emph{next level Intelligent Industrial Production Ecosphere} (IIP-Ecosphere) platform addresses, employing a highly configurable low-code based approach. 

In this paper, we introduce the design of this platform and discuss an early evaluation in terms of a demonstrator for AI-enabled visual quality inspection. This is complemented by insights and lessons learned during this early evaluation activity.

\end{abstract}
\section{Introduction}

%\begin{itemize}
%\item Artificial intelligence
%\item Artificial intelligence in production
%\item Training to Integration and Deployment
%\item survey \cite{iipSurvey}
%\item IIP-Ecosphere platform 
%\item Contents/structure of paper, specific contribution
%\end{itemize}

The field of artificial intelligence (AI) has made significant progress in recent years, in particular thanks to leveraging advances in the training of deep neural networks~\cite{lecun2015deep}. 
In \replaced[id=GP]{numerous}{various} areas AI performance is comparable to human performance, \eg for \replaced[id=GP]{object detection and image classification tasks}{the detection of objects in images}~\cite{pal2021deep,schnieders2019fully}. 
\added[id=GP]{As a result} AI-based methods are\deleted[id=GP]{, therefore,} increasingly being applied in a large variety of application domains for supporting automated decision processes~\cite{shinde2018review}.
An economically highly relevant application domain for AI is industrial production~\cite{angelopoulos2019tackling,peres2020industrial}. 
Here AI-based methods can increase effectiveness, improve quality and reduce costs as well as energy consumption~\cite{patalas2020ai,reimer2022identifying}. 
The most prominent example is AI-based condition monitoring, which intelligently reduces maintenance costs and down times~\cite{chen2021artificial}. \replaced[id=GP]{Further}{There are, however, also other} promising \added[id=GP]{application} areas \replaced[id=GP]{include}{such as} AI-based quality control~\cite{lee2019quality,palmer2020automated}, job-shop scheduling~\cite{denkena2021scalable,zhuang2018digital} and smart assembly systems~\cite{lin2022human}. 

A number of tools and libraries exist for easing the development and training of AI models – further augmented by the availability of powerful pre-trained models\cite{han2021pre}. 
\replaced[id=GP]{However}{in contrast to this}, other steps in the data science process for intelligent production are less well supported. 
This includes, for example, data acquisition (often not supported by the existing, long-lived and expensive legacy production systems) and the deployment of AI methods within close proximity of the target machines, \ie on industrial edge devices. 

%\textbf{AI $\rightarrow$ Deployment}

Operating an AI solution in an industrial context requires many supporting capabilities, including components for monitoring, resource management, data storage, integration with production and business processes, and configuration~\cite{SculleyHoltGolovin15}.
Although several Industrial Internet of Things (IIoT) platforms (also known as Industry 4.0 platforms) do exist, e.g., Siemens MindSphere, PTC ThingWorkx or AWS IoT, they usually fall short in some of the aforementioned aspects. Furthermore, they lack in flexibility, openness, freedom of installation (cloud vs. on-premise) or timely support for relevant standards~\cite{iipPlatfOver}, which are crucial for the demand-driven development and the operation of evolving, long-lived production systems. Researching and developing concepts for a flexible, open, standard-enabled and yet vendor-neutral IIoT platform is one of the core aims of the \IIPEcosphere{} project (next level \underline{I}ntelligent \underline{I}ndustrial \underline{P}roduction ecosphere)\footnote{\url{https://www.iip-ecosphere.de/}}. Our \IIPEcosphere{} platform aims for a low-code approach combining model-based configuration, code-generation techniques with relevant industrial standards, e.g. OPC UA (Companion Specs) or the currently trending Asset Administration Shells (AAS)~\cite{KannothHermannDamm+21,DetailsAAS}. An AAS follows an approach to describe assets, e.g., products or tools, in a machine-readable and vendor-independent manner, thus, smoothing information exchange and enabling innovative cross-company processes. 
%In AAS, an asset is modeled through properties and operations in a standardized manner.

%\textbf{More platform here?}

In this paper, we introduce the \IIPEcosphere{} platform for intelligent industrial production and an early validation in terms of a demonstrator use case on AI-enabled visual quality inspection\replaced[id=GP]{. We presented our use case}{, which was presented} at the \emph{Hannover Messe 2022}, one of the world's largest trade fairs\replaced[id=GP]{, exhibiting the state-of-the-art (SOTA) \wrt innovative technological developments in industry}{dedicated to technical development\comHE{improve}}. Besides technical aspects, the specific contribution of this paper is on the identification of challenges imposed by such an application and lessons learned for the engineering of intelligent IIoT platforms.  
The remainder of this paper is structured as follows. First, we discuss related work in Section \ref{sect_relatedWork} 
and then introduce the \IIPEcosphere{} platform in Section \ref{sect_platform}. 
In Section \ref{sect_useCase} we define our use case and then focus on the technical use case realization in Section \ref{sect_useCaseRealization}. 
In Section \ref{sect_lessonsLearned}, we discuss the lessons that we learned from this early demonstrator-based validation and in Section \ref{sect_futureConcl} we conclude with an outlook on future work.

%\comGP{1 Pages}

% ---------------------------------------------------------------------

\section{Related Work}\label{sect_relatedWork}

%A plethora of \textbf{IIoT platforms} exists. Some sources mention more than 450 different software platforms and \cite{platfIAO} found more than 1200 vendors. Moreover, the landscape changes often due to frequent mergers and dropouts~\cite{platfIAO}. 

In~\cite{iipPlatfOver}, we analyzed 21 industrial IIoT platforms. The platforms of highest industrial relevance were selected for this analysis from the large and evolving set of existing IIoT platforms - some sources mention more than 450 different software platforms and \cite{platfIAO} found more than 1200 vendors. We focused on 16 topics that are relevant for AI application in production, including AI capabilities, edge support, configurability, cloud vs. on-premise installation, use of standards including AAS integration. 
While most of the platforms provide some AI capabilities, the support is currently rather diverse, ranging from advanced service integrations to plain Python AI frameworks. Except for one platform, which allows on-premise installation, all other reviewed platforms are cloud-based.
Most of the analyzed platforms do not allow a deployment of own (AI) services on edge devices and recent standards such as OPC UA are rarely supported or, for AAS, not supported at all. 
As surveyed in~\cite{iipSurvey} with 75 companies, only 50\% utilize cloud, while only 30\% operate a platform. 
For the platform users, complexity, unclear license and cost models limit or even prevent the adoption of an IIoT platform. On the scientific side, \deleted[id=GP]{however, }an ongoing SLR reveals that researched platforms usually focus on a narrow inclusion of standards (MQTT, OPC UA), very limited configuration approaches, or non-industrial edge devices such as Raspberry PI. Notable platform concepts are discussed for protocols in~\cite{raileanu_edge_2018}, edge usage~\cite{chen_iot_2020,raileanu_edge_2018}, on-premise option, \cite{foukalas_cognitive_2020,lins_industry_2018,raileanu_edge_2018}, configurability~\cite{foukalas_cognitive_2020,lins_industry_2018} or AI capabilities~\cite{chen_iot_2020,foukalas_cognitive_2020,raileanu_edge_2018}.

In contrast, we present an AI-enabled IIoT platform that addresses identified gaps and industry-relevant capabilities. 
In particular, our platform focuses on openness, standard support, interoperability, configurability, industrial edge devices and production-relevant AI capabilities. 
Furthermore, our platform is to our knowledge the first to rely on a deep integration of AAS, e.g., to represent (runtime) component information or to steer platform operations.

\added[id=GP]{
With respect to the AI-services, we evaluate our platform using SOTA convolutional neural networks (CNNs),
which have enabled a significant number of recent breakthroughs
in the fields of computer vision and image processing~\cite{howard2017mobilenets}. 
CNNs' strength is their ability to automatically learn feature representations from raw image data\deleted[id=GP]{, alleviating the need for human guided feature extraction}~\cite{bansod2021analysis}.
With ever more advanced architectures emerging, CNNs are increasingly being utilized within an Industry 4.0 context -- from end-of-the-line visual inspection tasks \cite{palmer2020automated}
%, one-shot object detection for industrial robotics~\cite{schnieders2019fully}, 
to predictive maintenance and fault detection~\cite{silva2019assets}. 
In addition, while our current use case focuses on visual inspection\deleted[id=GP]{ using image data}, 
approaches where CNNs process encoded time-series data as images \deleted[id=GP]{(\eg log
Mel spectrograms~\cite{fedmodels})} are increasingly being used \replaced[id=GP]{in}{within the context of} intelligent manufacturing~\cite{HOANG201942,kiangala2020effective,WANG2019182}.% -- \eg through extracting log
%Mel spectrograms from sound data~\cite{fedmodels}.
%
\deleted[id=GP]{, being applied to:
predictive maintenance for conveyor motors using dual time-series imaging~\cite{kiangala2020effective};
defect detection and visualization in selective laser sintering~\cite{westphal2021machine};
fault recognition via image fusion of multi-vibration-signals~\cite{WANG2019182}; and
rolling element bearing fault diagnosis on vibration images~\cite{HOANG201942}.}}

% \comGP{1/2 of a page}

% \comGP{Related work on \textbf{AI} on anomaly detection, etc, focus on industry.}

% ---------------------------------------------------------------------

\section{IIoT Platform}\label{sect_platform}

% \comGP{2 Pages}

The \IIPEcosphere{} platform is planned as a 
\textit{virtual platform}, which can easily be added to existing production environments and augments existing functionality with AI capabilities, particularly through standardized protocols. During an intensive and interactive requirements collection for the platform~\cite{ESS22}, our industrial stakeholders expressed in particular three core demands, namely
\begin{enumerate} 
  \item the use of AAS to facilitate interoperability between products, processes and platforms.  While the standardization of formats for AAS is still in progress, the aim here is to explore the capabilities and limitations of AAS, \added[id=GP]{while} also considering more advanced uses, e.g., for software components. This early validation is also used to feed back requirements into the development of the AAS standard.
  %in particular to exchange data among devices and components of different vendors and to steer platform operations.
  \item the vendor-independent deployment of AI and data processing components within close proximity of the machines. This involves exploiting heterogeneous industrial edge devices in an open manner. In our case, this is smoothly integrated with the overall AAS approach of the platform by equipping the edge devices with a (vendor-provided or own) AAS providing deployment control operations. %While cloud-based execution is relevant as also identified in~\cite{iipSurvey}, our requirements clearly demand an on-premise variant. This limits the applicability of existing solutions such as Amazon Greengrass\footnote{\url{https://aws.amazon.com/de/greengrass/}} or IBM Edge Application Manager\footnote{\url{https://www.ibm.com/de-de/cloud/edge-application-manager}}.
  \item a loose integration of AI tools,
%   \comGP{regarding AI tools, services, etc, should we define somewhere above what we consider to be an AI service, and then stick to this term throughout the paper?}
%   \comGP{Of course, immediately after I write this comment, I see you have already defined what a service is below :-D.... still, perhaps better to use service(s) everywhere, and look for instances like tools, etc?}
   in particular for the development of an AI approach and for AI training. In other words, the platform shall support a data scientist in the development of the solution while not limiting the freedom of choice regarding AI and data science tools.
\end{enumerate}
%Currently, production control, SCADA or MES capabilities are out of scope. 

Based on more than 150 top-level requirements~\cite{ESS22}, we designed a layered, service-oriented architecture for our platform. As we have to deal with a wide variety of service realizations including Java classes, Python scripts and even binary executables, we employ a rather open notion of the term \textit{service}\footnote{In more details, for us a service is a function with defined input- and output data/management interfaces, which can process data either in synchronous or asynchronous fashion, and can optionally be distributable.}. Services can be application-specific or generic, adaptable ones shipped with the platform. 

\begin{figure}[htbp] % trim={<left> <lower> <right> <upper>}%\fbox{
\centerline{\includegraphics[trim={1.5cm 4.5cm 1cm 5.5cm},clip, scale=0.53]{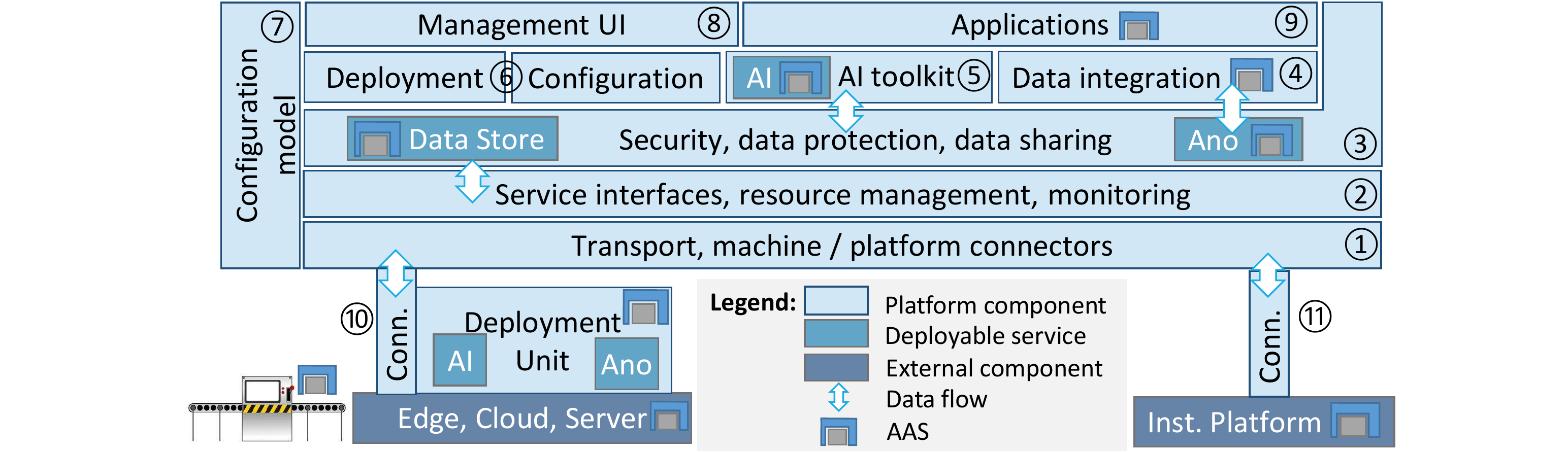}}
\caption{Platform component overview.}
\label{fig_platform}
\end{figure}

Figure \ref{fig_platform} illustrates the main building blocks. The lowest layer \textcircled{1} focuses on the communication, i.e., the data \textit{transport} among components (such as MQTT or AMQP), but also with machines or other platforms through pluggable \textit{connectors} (e.g., OPC UA, MQTT, AAS). 

The \textit{services layer} \textcircled{2} defines the service management interfaces and realizes mechanisms for resource management and monitoring. Moreover, this component contains language-specific execution environments to ease the realization, integration and execution of individual services, e.g., for Java or Python. 

The \textit{security services layer} \textcircled{3} realizes additional platform-provided functionality that allows for enhancing security \replaced[id=GP]{through}{and} secure data processing and data sharing. Here, one example service is a generic anonymization and pseudonymization service that can safeguard the processing of data about persons. 

The components in the next layer realize advanced services. For the data processing, these are a semantic \textit{data integration} \textcircled{4} of multiple data sources as well as production-specific AI methods in terms of an \textit{AI toolkit} \textcircled{5}. Besides an integration of open source components such as Python packages, the platform can also integrate commercial components, e.g., the RapidMiner Real Time Scoring Agent (RTSA)\footnote{\url{https://docs.rapidminer.com/9.4/scoring-agent/deploy-rts/}}, a generic execution environment for AI workflows. In the same layer, two core components of the platform are located, the mechanisms for consistent configuration and control of the heterogeneous service deployment to execution resources  \textcircled{6}. The configuration component is based on a cross-cutting configuration model \textcircled{7}, which integrates classical software product line concepts~\cite{LSR07} like optionalities, alternatives and constraints with topological  capabilities~\cite{EichelbergerQinSizonenko+16} for modeling the graph-based data flows of an IIoT application. 

The top-most layer contains the integrated applications as well as a \textit{web-based user interface} \textcircled{8} for managing the platform. An \textit{application} \textcircled{9} consists of contributions to the configuration model defining the data types and generic or application-specific services to be used as well as their data flows.

Depending on the application, connectors and services can be distributed to the executing resources \textcircled{10}, e.g., edge devices in virtualized manner, i.e., as (Docker) containers. Therefore, a part of the platform is running on the device, which executes deployment and service commands issued by the central management parts of the platform. Moreover, the platform can communicate with already installed software or platforms via connectors \textcircled{11}. Many tasks of creating such an application are automated in a model-based, generative fashion based on the configuration model. Examples are the generation of interfaces and basic implementations of application-specific services, the adaptation of generic services to application-specific data formats by glue code, the generation of machine connectors based on low-code specifications in the configuration model, or integration and packaging of the application.

As mentioned above, the platform supports actual standards in the IIoT domain and, in particular, the currently evolving AAS. As indicated in Figure~\ref{fig_platform}, components that are provided by third parties such as services, applications or devices/machines are described in terms of an own AAS, e.g., through a (pragmatic use of) the nameplate for industrial devices~\cite{genFrame11}. The platform also maintains an encompassing AAS for itself. This consists of 11 submodels, which references the third-party AAS and provides additional elements, e.g., runtime monitoring properties~\cite{aasMonitoring21} or management operations. 

For the development of the open source \IIPEcosphere{} platform\footnote{\url{https://github.com/iip-ecosphere}}, we build upon and integrate at the time of writing 23 open source components, in particular from the Eclipse IoT ecosystem such as BaSyx\footnote{\url{https://www.eclipse.org/basyx/}} for the realization of AAS. All integrated components (including BaSyx) are treated as optionalities or alternatives~\cite{LSR07} and utilized in the platform only through platform interfaces and code adapting the component to the respective interface. On the one \replaced[id=GP]{hand}{side}, this facilitates openness and systematic variability, e.g., of the platform transport protocol. On the other \replaced[id=GP]{hand}{side}, an interface-based integration eases coping with unexpected external changes of components, as in most situations only the implementation of the interface must be adjusted accordingly.

%platform handbook \cite{iipPlatf}

% ---------------------------------------------------------------------

\section{Demonstrator Use-case Description}\label{sect_useCase}

%\comGP{Description: 1/2 of a page}
For a first validation of our platform we have selected a use case consisting of a simple AI-based quality inspection for individualized products. \replaced[id=GP]{T}{Although of small scale t}he use case utilizes industrial components such as a cobot and an industrial edge device for realistic validation results. \replaced[id=GP]{Therefore, despite being of small scale}{In more detail}, the use case is tailored to investigate the following questions:
\begin{enumerate}[start=1,label={Q\arabic*}]
    \item\label{qAI2edge} What are the practical challenges when bringing AI-based quality inspection close to the machines, i.e., onto an edge device? In more detail, What is the ability of the platform to automatically deploy AI services on the edge irrespective of model specific dependencies?
    \item\label{qAAS} How can AAS be best exploited in support of quality inspection in lot-size-one (individualized) production settings?
    \item\label{qIntegrateTech} How well does the platform support integrating different involved technologies (AI, cobot, edge, etc.) and developing the respective IIoT application?
    \item\label{qInter} How can an IIoT application be developed by an interdisciplinary team as required in such a setting?
    %, here consisting of data analysis, automation and factory experts and software engineers, on the development of the demonstrator. 
    Often, interdisciplinarity is seen as a special challenge when realizing, e.g., data analysis~\cite{HEG+18} or cyber-physical systems~\cite{FeichtingerMeixnerRinker+22}.
\end{enumerate}

As products we are using small aluminum cars, which are produced for this purpose. The cars vary \wrt product properties such as wheel color or engravings. Accompanying production, AAS capturing those properties are generated and made available to the platform via an AAS registry. 
Quality inspection is performed based on three position images taken of the car models using a 5 axis cobot arm with a mounted camera. 
AI models trained for this purpose are used for quality control and deployed onto the edge for this purpose. 
Based on the AI results, the quality is measured along two axis: a) conformance with production specification (as given by the product's AAS) and b) quality of production (existence of scratches). 

%\comGP{1 page}

%In more details, we validate the following capabilities of our  platform:

%\begin{itemize}
%    \item Transferability or the ability to cope with non-stationary domains, \eg being deployed in a new environment at inference time.
    %\item Switching AI services \say{on-the-fly}.
%    \item Rapid deployment of new AI models and IIoT applications. 
%\end{itemize}

\section{Demonstrator Realization}\label{sect_useCaseRealization}

The demonstrator use case has been sucessfully implemented. In the following, we provide insights into the employed hardware, the interaction of the components/services, the role of our platform and describe the AI service. 

The IIoT hardware of the demonstrator consists of a Universal Robots UR5e cobot\footnote{\url{https://www.universal-robots.com/de/produkte/ur5-roboter/}} equipped with a Robotiq wrist cam\footnote{\url{https://www.universal-robots.com/plus/products/robotiq/robotiq-wrist-camera/}} as well as a Phoenix Contact AXC 3152, a combined Programmable Logic Controller (PLC) and edge device\footnote{\url{https://www.phoenixcontact.com/de-de/produkte/steuerung-axc-f-3152-1069208}} with \added[id=GP]{an} Intel Atom processor, 2 GByte RAM, and 32 GByte SD-based hard drive. In addition, a usual PC plays the role of the central IT running our platform and a tablet is used to present the application user interface.

The software side consists of four main components, the \texttt{Cam source} as image input, the \texttt{Python AI}, the \texttt{Action Decider} controlling the overall process and the \texttt{App AAS} as data sink. Furthermore, two connectors, thr \texttt{OPC UA connector} and the \texttt{AAS connector} integrate the hardware as well as the external car AAS.

\begin{figure}[htbp]% trim={<left> <lower> <right> <upper>}\fbox{
\centerline{\includegraphics[trim={2.7cm 4.5cm 0cm 5cm},clip, scale=0.48]{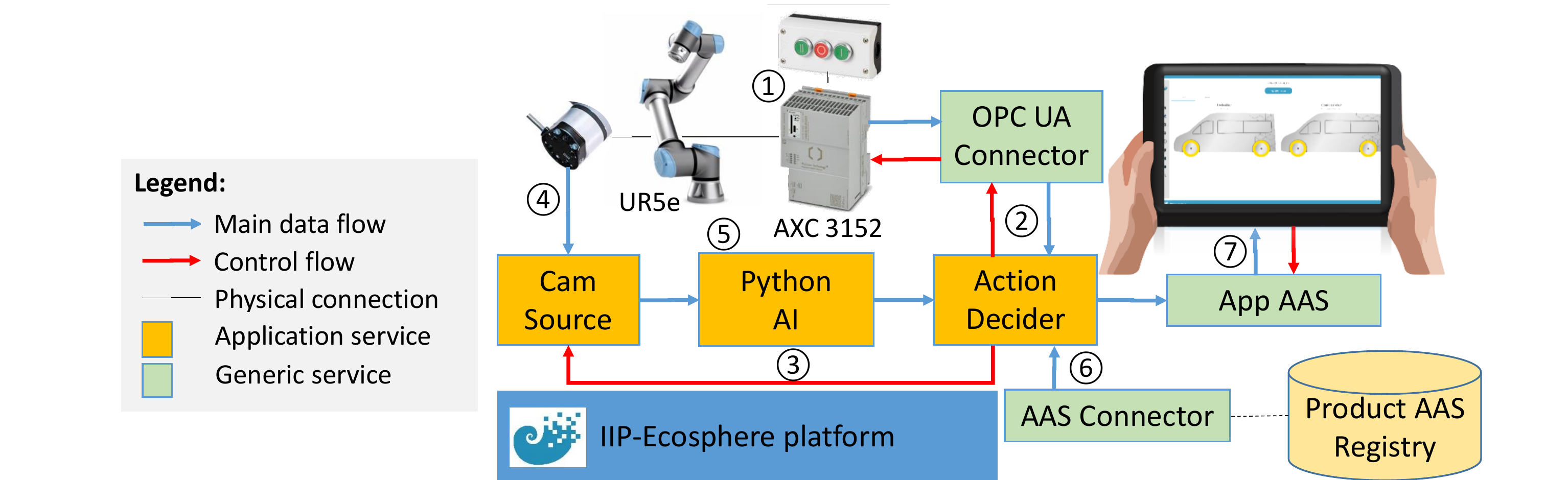}}
\caption{Demonstrator components and data flows.}
\label{fig_demoOverview}
\end{figure}

As illustrated in Figure \ref{fig_demoOverview}, those components together form the demonstrator application\footnote{We plan to make the source code of the demonstrator available as open source. 
% \comHE{Add a link to a short video?} 
% \comGP{Good idea. Do you have one? (I have one short one from the messe, where the cobot enters the first inspection position. Sadly not the full routine.} \comGP{Have just messaged Trong to see if he has one. He has the following \url{https://drive.google.com/file/d/1-70SgB1jjd1telhjp1CNy2Cd-5dZDHq-/view?usp=drivesdk}. Bit messy...)}
% \comGP{Just thinking. Annika was also taking videos of stuff at the Klausrtagung. Perhpas she also captured the cobot demo. What about \url{https://sync.academiccloud.de/index.php/s/NNPSbA846SyqTKd}}
}. We briefly describe the design of the components and their interactions in terms of a normal execution sequence\footnote{A video showing our setup is available at: \url{https://youtu.be/36Xtw1L2XkQ}.}. 

The quality inspection process is initiated by pressing a physical button connected to the AXC \textcircled{1}, which reflects the button status on its OPC UA server. A customized OPC UA connector provided by our platform observes the change and informs the \texttt{Action Decider}. Alternatively, the user can press a UI button on the tablet. The UI is based on an application AAS provided by the \texttt{App AAS} service, which provides processing results as well as operations, such as starting the inspection process.

The \texttt{Action Decider} \textcircled{2} initiates a movement of the robot arm to one of the three visual scanning positions (QR position, left or right car side) and requests the \texttt{Cam Source} \textcircled{3} to take a picture from the wrist cam \textcircled{4}. In the QR position, the \texttt{Cam Source} performs a QR-code detection and augments the input data by the QR payload. Pictures are subsequently taken for the left and right car side. Each picture is streamed to the \texttt{Python AI}, which aims to detect the product properties \textcircled{5}. The AI results including the actual picture form the input to the \texttt{Action Decider}. Based on the QR payload, the \texttt{Action Decider} obtains the product AAS \textcircled{6} and compares the information detected by the AI and the configured information. 

The comparison results as well as trace information on all intermediary steps are handed over to the \texttt{App AAS} \textcircled{7}, which is presented by the Angular-based application UI on the tablet. The UI application is entirely based on AAS. It enables following the entire quality inspection process, visually browsing backward and forward on demand. Furthermore, is confronts in an illustrative way the picture of the product according to the specification with a picture of the product as it is seen by the AI. This has proven to be very useful for the demonstration of the use case on the trade fair. Moreover, it allows the inspection of all underlying services running on the platform through their individual AAS.

With respect to the platform, the application as depicted in Figure \ref{fig_demoOverview} is modeled as data flow in the platform configuration model. The application instantiation process created application-specific interfaces of the three application-specific services (\texttt{Cam Source}, \texttt{Python AI}, \texttt{Action Decider}), customized generic connectors and services (\texttt{OPC UA connector}, \texttt{AAS connector} and \texttt{App AAS}) for their use in the application, generated glue code for the distributed stream-based service execution and performed the Maven-based integration and packaging of the application. In summary, the application (except for the application UI) consists of about 8.2 KLOC, whereby 51\% are generated and 49\% are manually written (30\% production and 19\% testing code). At runtime, the platform cares in particular for the distribution of the services to the target devices, the execution of services and container as well as for the runtime monitoring.

Below, we detail \deleted[id=GP]{now} the most complex component of the application, the Python-based \replaced[id=GP]{AI}{artificial intelligence} service.
%\comGP{2 pages}
%\subsection{Data sources: Images, OPC-UA, AAS}
%\comHE{TBD}
%\subsection{AI Components}
%\comGP{1/2 page}
% \comGP{@Hieu, Sofiane: Can you guys paste in: 1.) preprocessing steps, the edge detection stuff; 2.) rough details of the AI components, including model architectures, training details, hyperparams, etc; 3.) Description of the color thresholding that we ended up going with; and 4.) Perhaps model sizes, etc? Doesn't have to be pretty, if you can just paste the details in as bullet points, then I can polish it into paragraphs, tables, etc.}
Our use-case presents us with four AI-related challenges: 
a)~tire color detection; %(straight forward, no need AI model for this task)
b)~detection of an engraving towards the rear of the vehicle, if present, then both on right rear and left rear sides;
c)~classification of the number windows $[1, 3]$;
and d)~the detection of scratches (drawn on using a black non-permanent marker pen\added[id=HE]{, also allowing for an interaction opportunity with the audience}). 
Task a) is trivially solved using a range based color threshold approach, where ranges for each of the four color types (red, green, yellow, black) are found upon converting our training images into HSV color-space.
For the remaining tasks we utilized deep \replaced[id=GP]{CNNs}{neural networks}.
We find that each problem benefits from a custom preprocessing step, prior to feeding the image into the model. %\comHE{following 3 words needed?} for model to training / predicting. 
Furthermore, given the limited capacity of the edge device, we train a distinct model for each task that can be loaded when required, instead of training one big model for all three tasks.
Given the small size of our dataset (consisting of 200 images), common dataset augmentation techniques (\eg random flipping, \deleted[id=GP]{random} rotation and \deleted[id=GP]{random} zooming in/out) are utilized to \deleted[id=GP]{help increasing the size of dataset and} 
avoid overfitting our models.
We describe each model and respective preprocessing steps in detail below\replaced[id=GP]{. Hyperparameters that are consistent across our models include: an Adam optimizer; a learning rate of 1e-4; $\beta_1=0.9$; $\beta_2=0.999$; $\epsilon=$1e-07; $1000$ training epochs; a batch size of $20$; Relu activation layers; and a training / test ratio of 9:1}{, and list the hyperparameters that were consistent across our models in Table~\ref{tab:hyperparams}}.
%
%{\textbf{Results:}} 
%
All models reach $\sim99\%$ and $\sim95\%$ accuracy on the training and testing datasets respectively.

%In each task we would have 1 model alongside with it and also each task we use different filter/mask for the preprocessing and then put it to the model to train/predict.
%Given our small dataset we use augumentation (random flipping, random rotation and random zooming in/out) to increase the datasize and avoid overfitting of the network.
%Preprocessing :

\begin{figure}
\centering
\begin{subfigure}[b]{0.24\textwidth}
\rotatebox[origin=c]{180}{\includegraphics[width=\textwidth]{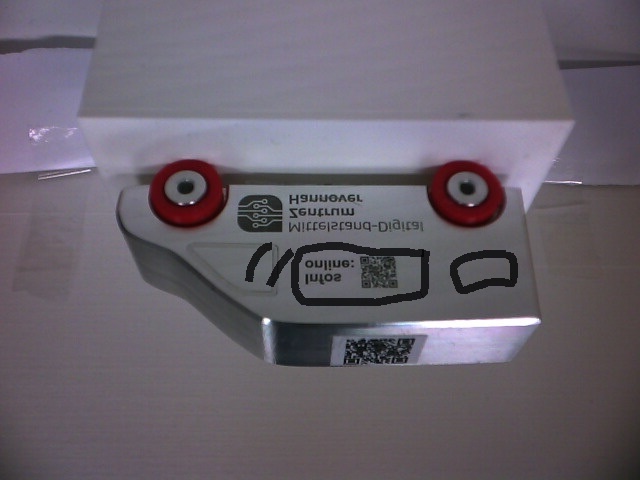}}
\end{subfigure}
\hfill
\begin{subfigure}[b]{0.24\textwidth}
\rotatebox[origin=c]{180}{\includegraphics[width=\textwidth]{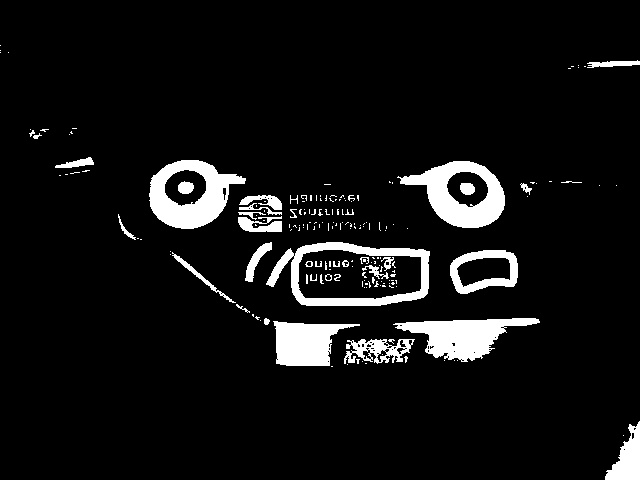}}
\end{subfigure}
\hfill
\begin{subfigure}[b]{0.24\textwidth}
\rotatebox[origin=c]{180}{\includegraphics[width=\textwidth]{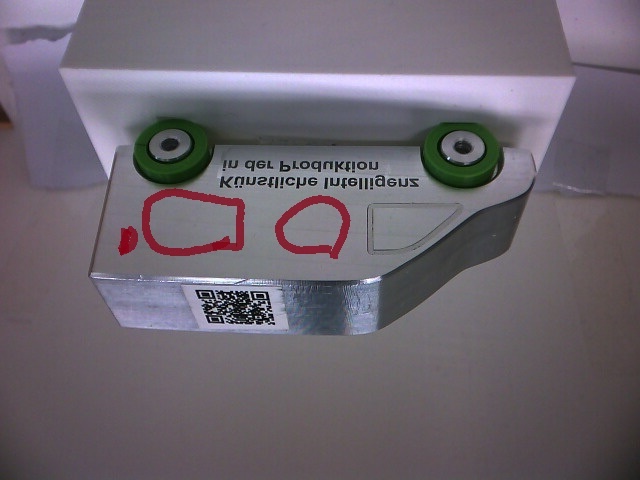}}
\end{subfigure}
\hfill
\begin{subfigure}[b]{0.24\textwidth}
\rotatebox[origin=c]{180}{\includegraphics[width=\textwidth]{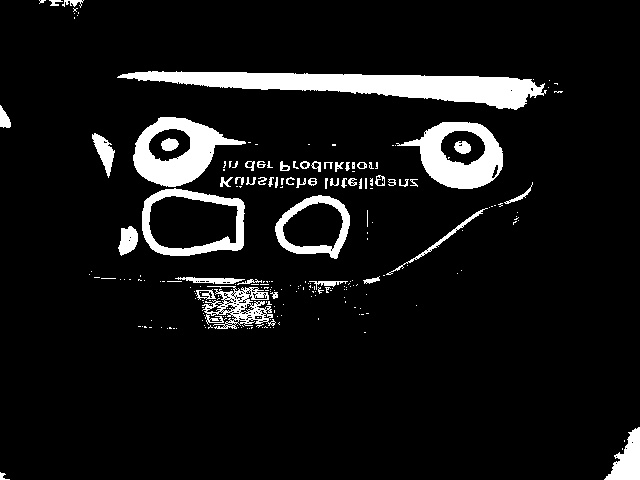}}
\end{subfigure}
\caption{Result of applying our preprocessing prior to scratch detection. }
\label{fig:scratch_detection}
\end{figure}

{\textit{Scratches detection:}} For preprocessing we
first apply the same color range based thresholding approach to
%obtain a mask to 
eliminate colors within the range of the car's color\deleted[id=GP]{, thereby highlighting the colors that are outside the car's color range} (see Figure~\ref{fig:scratch_detection}).
The CNN for this task consists of separable convolutional layers~\cite{howard2017mobilenets}, representing a light weight alternative to normal convolutional layers without significantly impacting performance. 
\deleted[id=GP]{In-fact, Howard et al.~\cite{howard2017mobilenets} found  that MobileNet (3.3M parameters) only performed slightly worse than Inception-V3 (23.2M parameters) achieving an accuracy of 83.3\%  vs 84\% on the Stanford Dogs dataset.} 
Separable convolutional layers are therefore well suited for edge devices with limited capacity \wrt memory.
The model is trained using a binary cross entropy loss function (sigmoid).

\begin{figure}
\centering
\begin{subfigure}[b]{0.24\textwidth}
\rotatebox[origin=c]{180}{\includegraphics[width=\textwidth]{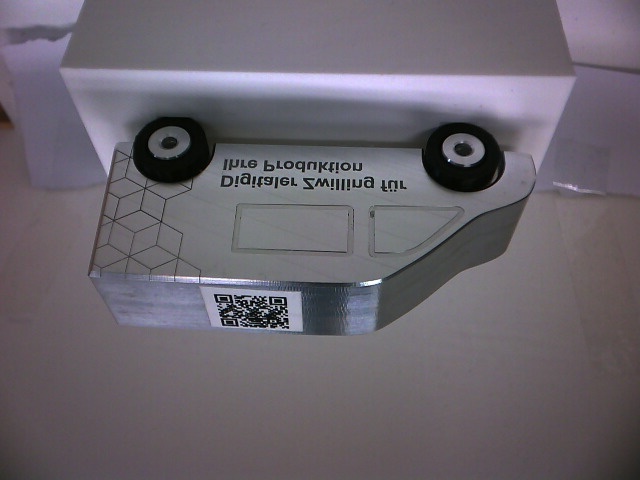}}
\end{subfigure}
\hfill
\begin{subfigure}[b]{0.24\textwidth}
\rotatebox[origin=c]{180}{\includegraphics[width=\textwidth]{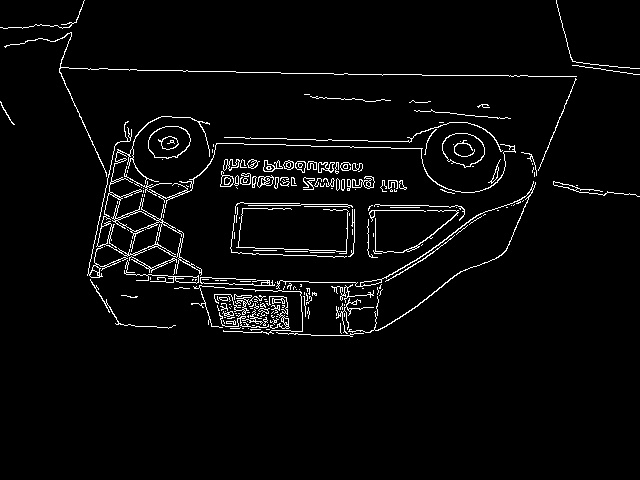}}
\end{subfigure}
\hfill
\begin{subfigure}[b]{0.24\textwidth}
\rotatebox[origin=c]{180}{\includegraphics[width=\textwidth]{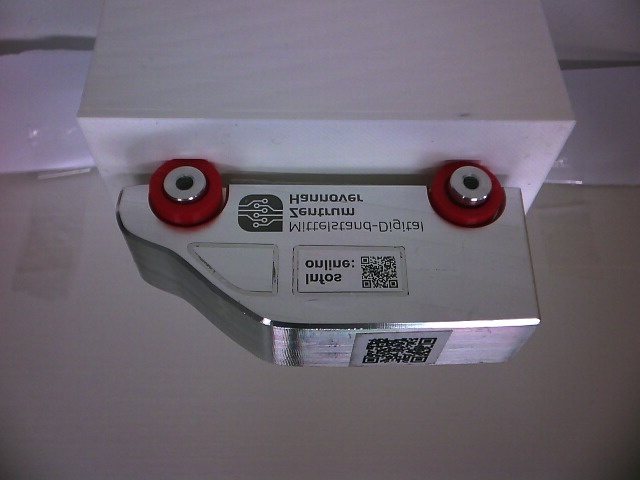}}
\end{subfigure}
\hfill
\begin{subfigure}[b]{0.24\textwidth}
\rotatebox[origin=c]{180}{\includegraphics[width=\textwidth]{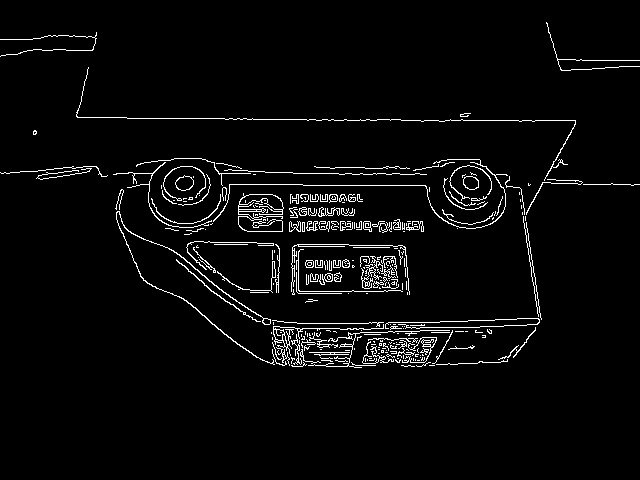}}
\end{subfigure}
\caption{Two examples of before/after preprocessing for detecting the engraving.}
\label{fig:engraving_detection}
\end{figure}

{\textit{Engraving detection:}} For this task we used OpenCV's edge detection to find a mask that can highlight the engraving (see Figure~\ref{fig:engraving_detection} for examples). 
The rest of the process is the same as for the scratch detection outlined above.

\begin{figure}
\centering
\begin{subfigure}[b]{0.3\textwidth}
\rotatebox[origin=c]{180}{\includegraphics[width=\textwidth]{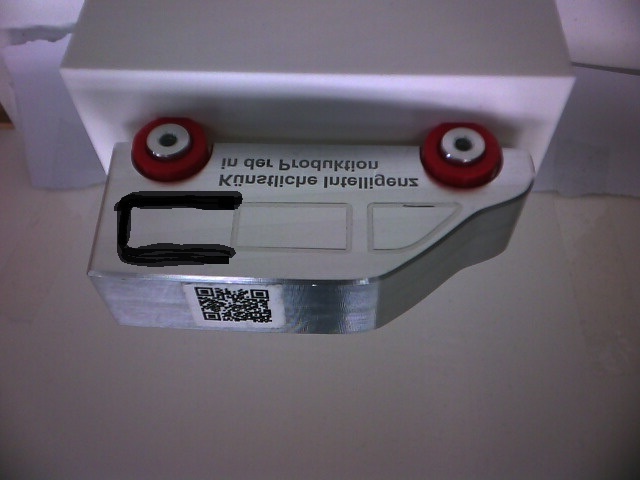}}
\caption{Original Image}
\end{subfigure}
\hfill
\begin{subfigure}[b]{0.3\textwidth}
\rotatebox[origin=c]{180}{\includegraphics[width=\textwidth]{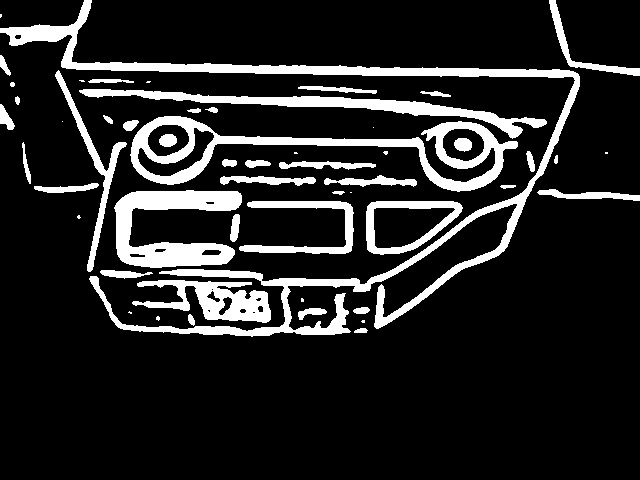}}
\caption{Mask 1}
\label{fig:window_detection:M1}
\end{subfigure}
\hfill
\begin{subfigure}[b]{0.3\textwidth}
\rotatebox[origin=c]{180}{\includegraphics[width=\textwidth]{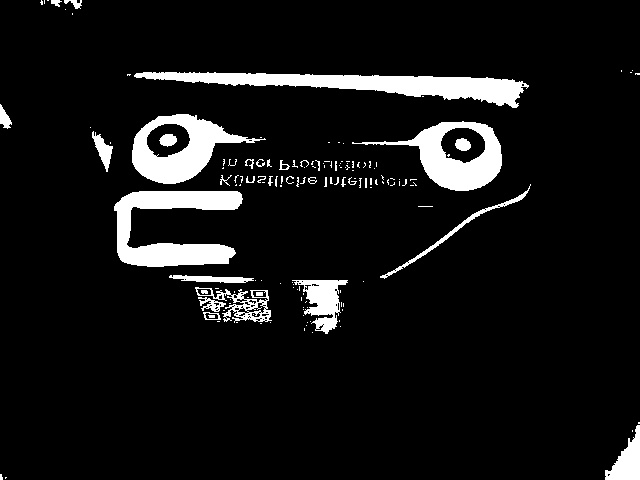}}
\caption{Mask 2}
\label{fig:window_detection:M2}
\end{subfigure}
\caption{Window detection task example, where a window is drawn
onto the vehicle. Both mask from Sub-Figures \ref{fig:window_detection:M1} and \ref{fig:window_detection:M2} are fed into the model, the first highlighting both windows and {scratches}, while Mask 2 emphasises the {scratch}.}
\label{fig:window_detection}
\end{figure}

{\textit{Windows detection:}} 
This task is more challenging than the others, as windows 
can be confused with \say{scratches} (\deleted[id=GP]{or rather, individuals
using the marker pen to drawn on windows, }see Figure~\ref{fig:window_detection}\deleted[id=GP]{ for an example}).
To overcome this challenge we utilize a model 
that receives two inputs: 1.) a mask that highlights both the car windows and the scratches; and 2.) the mask from the first task (scratches detection). 
Feeding in both masks allows the model to learn to recognise anomalies.
\deleted[id=GP]{The CNN for this task is larger than the models from the \replaced[id=GP]{other}{engraving and scratch detection} tasks due to the two inputs.} 
%
%Hyperparameters: Adam optimizer; learning rate=0.0001; beta 1=0.9; beta 2=0.999; epsilon=1e-07; Relu for the activation layers; and 
A categorical cross entropy loss function (softmax) is used for training the model, since we have \deleted[id=GP]{3 labels (}three different window configurations\deleted{)}.

To gain first insights into the performance of executing the AI on the employed hardware, we pragmatically measure the inference time per image in the target setup. Per image, the \texttt{Python AI} consumes \deleted[id=GP]{per image} 0.4-0.5~s on the PC and 3-5~s on the edge device. A performance drop is not surprising here, as the edge is a resource-constrained device, which dedicates parts of its CPU and memory to hard realtime processes. Moreover, the hard drive is SD-based, which may limit file transfer, e.g., when reading AI models. However, the measurements of both devices exceed by far the required production machine pace of 8~ms~\cite{ESS22}. Thus, further optimizations are needed, e.g., by employing a GPU or TPU co-processor (a TPU module is available for the AXC 3152) or programming level optimizations, e.g., to avoid re-loading models in a streaming setup.

\deleted[id=GP]{
\begin{table*}[]
    \centering
    \caption{Hyperparameters used by our neural networks.}
    \begin{tabular}{p{1.5cm}|p{1.3cm}|p{0.7cm}|p{0.8cm}|p{0.8cm}|p{1.1cm}|p{1cm}|p{1.5cm}|p{1.6cm}}
    \toprule
         Optimizer & Learning Rate & $\beta_1$ & $\beta_2$ & $\epsilon$ & Epochs & Batch Size  & Activation Layers & Training / Test Ratio  \\
        \hline
         Adam &  1e-4 & 0.9 & 0.999 & 1e-07 & 1000 & 20 & Relu & 9:1 \\
        %  Optimizer & Adam \\
        %  \hline
        %  Epochs & \comGP{Sofiane is checking with Hieu} \\
        %  \hline
        %  Batch size & \comGP{Sofiane is checking with Hieu} \\
        %  \hline
        %  Early stopping? & \comGP{Sofiane is checking with Hieu} \\
        %  \hline
        %  Learning Rate & 0.0001 \\
        %  \hline
        %  beta 1 & 0.9 \\
        %  \hline
        %  beta 2 & 0.999 \\
        %  \hline
        %  epsilon & 1e-07 \\
        %  \hline
        %  Activation layers & Relu \\
        %  \hline
        %  Training and Testing Dataset Splits & 9:1 ratio \\
    \bottomrule
    \end{tabular}
    \label{tab:hyperparams}
\end{table*}}

% ---------------------------------------------------------------------
 \input{LessonsLearned}

\section{Future work \& Conclusion}\label{sect_futureConcl}

%Artificial Intelligence is a key technology to realize the vision of intelligent industrial production and Industry 4.0. Realizing AI-based IIoT applications and deploying them successfully in industrial setting deserves a significant amount of supporting services and technologies, usually offered in terms of IIoT software platforms. While current solutions offer stability and a plethora of (legacy) integration forms, they often limit the decision space of the user and fall short in openness, vendor-neutrality, standard support or systematic AI integration. One mission of the \IIPEcosphere{} project is to research concepts addressing these gaps and to demonstrate them in terms of an open source IIoT platform.

In this paper, we provided an overview of our IIoT platform approach -- the open source \IIPEcosphere{} platform -- and introduced its architecture. While designing and developing such a platform is a technical challenge in itself, although supported through more than 20 integrated open source components, it is important to validate and evaluate the approach with industrial devices and use cases. As a first step in this direction, we presented an AI-based visual quality inspection, which combines industrial devices (edge, cobot), recent and upcoming standards (OPC UA, AAS) with actual AI methods. From the realization of the demonstrator, we derived and discussed lessons learned, which pinpoint actual problems but also indicate positive experiences that can be helpful for other work in the field. 

As future work, we plan to investigate some of the open problems from our lessons learned discussion and further applications, e.g., selection and optimization of AI for resource-constrained devices, federated AI models \cite{fedmodels} architecture in a federated multi-device setup, automated container building for Industry 4.0 applications or AAS-based plug-and-play for industrial edge devices during onboarding into an IIoT platform. Besides these topics, we will improve and stabilize the realization of our platform, which already raised various industrial interests outside the consortium, as a basis for deeper evaluations and further demonstrators.

%\comGP{Whatever remains....}

\bibliographystyle{splncs04}
\bibliography{sasi4}
\end{document}

%% file: LessonsLearned.tex
\section{Lessons Learned}\label{sect_lessonsLearned}

%\comGP{2-3 pages}\comHE{decide on sequence}

We implemented our use case and successfully demonstrated it and the underlying platform at \added[id=GP]{Hannover Messe 2022 --} a large international industrial trade fair. In addition, in realizing the use case we collected a number of experiences and encountered a diverse set of unexpected issues. 

In this section, we summarize and discuss the most relevant experiences and issues as lessons learned for the target community. In this context, we revisit also the questions Q1-Q4 from Section \ref{sect_useCase}. 

%
%\comGP{I wonder if we need to provide some context here. In particular, when we make claims about lack of documentation, kind of invites a reviewer to say "Hey, I have you seen the following url: blah.com.!?". Should we frame this as essentially a six week hackathon, and state when we say lack of sufficient documentation, we mean that we were unable to find any, and it is therefore wrt the best of our knowledge? Or perhaps we do need to be bold, and indeed state that to-date there is a lack of documentation for x.}
%the topics \comHE{IIoT integration, dataflow component testing, contaier virtualization, software versioning, use of AI, timing issues and AAS}.

%Although we showcased our demonstrator and the underlying platform successfully in an international fair context, we also expected different unexpected issues and road blockers. In this section, we summarize the most relevant issues for the target community and discuss lessons learned for the topics \comHE{IIoT integration, dataflow component testing, contaier virtualization, software versioning, use of AI, timing issues and AAS}.

\textbf{IIoT integration}: Incorporating the cobot, the edge, the camera as well as the IT components and the AI models into an IIoT application (and make them interact \say{smoothly}) was well supported by the platform. This is especially due to integrated protocols, generated connectors as well as data flows (\ref{qIntegrateTech}). However, it also proved to be much more challenging than expected. A major reason for this is that several technological borders have to be bridged, where each of those technologies comes with their own paradigms, languages, and (only partially documented) limitations. We especially experienced this, when incorporating the cobot (edge - cobot interaction) and the camera (edge - camera interaction). 
%Setting up the IIoT hardware (PLC, cobot and camera) caused several unexpected issues. First, we discuss the PLC-cobot setup then the PLC-camera integration.

For the \textit{edge-cobot communication}, several integration paths have been explored. This included ProfiNet, direct wiring and a software solution based on the robot operating system ROS\footnote{\url{https://www.ros.org/}}. Although ProfiNet is a standardized network protocol and the Phoenix Contact PLCnext programming environment provides a function brick for the integration of UR cobots, the establishment of a reliable bi-directional communication between edge and cobot was hindered by a lack of sufficient documentation. Alternatively, direct wiring of digital in- and outputs between edge and cobot is a feasible approach, but limited in flexibility due to the number of supported wires and difficult in realization, as here low-level signal details must be considerd for the programming. In the end a ROS-based solution was selected due to its easier integration with the platform.

%we initially relied on ProfiNet, a standardized field-level protocol. For easing the integration, the Phoenix Contact PLCnext environment provides a function brick for UR cobots, which can be configured via GSDML\footnote{\url{https://www.feldbusse.de/Profinet/gsd.shtml}} for the specific robot at hands. Based on this brick, we realized a state-based PLC program that can be triggered by the platform via OPC UA, i.e., the platform requests a part of the demonstrator sequence and the PLC coordinates one or multiple robot movements and takes care of safety incidents. However, depending on the actual operation mode of the robot (local, remote), parts of the ProfiNet protocols seem to be disabled. As a consequence, reading values from the robot worked while sending values and commands partially failed. Thereby, integer values such as the requested movement number were acknowledged by the robot in reversed byte order. We were not able to figure out the root cause in the given time, because the robot control tablet does not display the actual variable values and, moreover, the robot manual does not clearly state the impact of the modes on ProfiNet. As first fallback, we connected PLC and cobot by a multi-wire cable, one wire per digital signal. Although this approach worked, we finally decided due to time limitation to control the robot by software via ROS. 

From the \textit{integration of the camera with the edge} we learned that the capacity of the involved devices might depend upon the way they are connected - and not all devices support all types of connections. For the case of the camera, for example, some functionalities such as explicitly controlling the focal length, lighting, or image preprocessing - although supported when connected via USB - are not accessible via other communication paths. Since the employed edge does not support USB, it had to be connected indirectly via a Web Server and networking limiting the available functionalities. 
  
%The \textit{camera} is physically connected via USB. However, the PLC and the camera cannot be connected directly, as, in contrast to some industrial PCs, the AXC does not support USB. The robot hardware can serve the camera via a webserver, i.e., the PLC can access the camera via Ethernet. However, control over the camera via webserver is limited, i.e., we were not able to explicitly control the focus length or to trigger advanced functions, e.g., image preprocessing such as edge detection. An alternative is to connect the cam via USB to the server, where e.g., a Python library can address the advanced camera functions. However, our aim was to execute the cam source and the data preprocessing on the Edge device rather than on the server, so we did not realize this alternative.
In summary, we learned that integrating IIoT devices requires cross-domain expertise (\ref{qInter}), e.g., in our case from the automation domain, the robot manufacturer and the application/software side. Here, more plug-and-play opportunities or agreed protocols such as OPC UA are desirable. However, this seems to require further standardization efforts as the actual OPC companion specification for robots only allows for reading the robot state rather than taking control over a robot. 

\textbf{Container virtualization on constrained resources:} To support heterogeneous devices, the platform encourages the virtualization of the distributed service execution through container technology. However, building a container for an edge device requires a careful software selection, consideration of processor capacities, memory and disk spaces as well as subsequent optimization of the container  (\ref{qAI2edge}). For achieving the container-based AI deployment on the employed edge, for example, several iterations were required. The initially used TensorFlow is shipped as a monolithic 2 GByte Python package.
However, installing all required components results in a container with a 4.2 GByte disk footprint, which can easily exceed the capacity of the employed edge device. Furthermore, TensorFlow did not support the Atom processor of the edge device. We therefore switched to TensorFlow light, which requires an upfront conversion of the AI model and, in effect, leads to a container size of about 2.6 GB. 
An alternative is to utilize tailored compilations for such devices, if available. Here, better modularization of such libraries, e.g., an inference-only TensorFlow as well as official, easy-accessible builds of native software for different processor versions would be desirable. However, inference only models are at a disadvantage within non-stationary domains that require online learning.

\textbf{AI in Production Context:} Our experience regarding AI for production is twofold. On the one hand, it is important to carefully assess whether complex AI methods really bring benefits over algorithmic methods, e.g., a threshold-based method for wheel color detection. This is in particular important when deploying AI to resource constrained devices. On the other hand, AI tasks in an industrial production context, are in many cases more complex than expected from theory, because the physical environment of the AI application may have a strong influence, e.g., light, heat, vibration. In our case, AI needed to be tuned due to light reflection on polished aluminium and changing lighting conditions in different settings (lab, trade fair ground). Furthermore, careful data preprocessing also proved to be very effective already in our relatively controlled environment.

\textbf{Service Testing Sandbox:} Besides platform components, it is essential to test application-specific services and components. Setting up such tests is a complex task, depending on the frameworks, techniques and conventions employed by the platform. We experienced this when integrating the Python-based AI service as it implicitly involves different programming languages as well as the underlying stream processing environment, in our case Spring Cloud Stream (SCS)\footnote{\url{https://spring.io/projects/spring-cloud-stream}}. In other words, such a test requires setting up a SCS-based service test, which calls the managing Java stream component and, through the Python service environment, the actual Python service code. Moreover, it is desirable to enable testing for application developers, who do not (yet) have deep knowledge about platform internals (\ref{qInter}). To ease testing for platform users and based on the integration experiences that we made, we plan to generate a test basis for each application service, which transparently sets up the SCS environment and also feeds the service with test data in a uniform manner.

%While we tested the generated connectors and most of the Java-based services, we skipped testing the Python AI service, because setting up a test environment for such a SCS-based service is a rather complex task. However, while preparing the Python part for integration (the platform expects a class rather than a script), several bugs were introduced accidentally. Identifying these bugs as part of the application integration tests was very time consuming as for each modification, the application build process had to be executed, which runs for several minutes. We learned here that testing facilities for developers with different technical background, e.g., data analysts with limited Java background, are essential for a platform. We plan to generate for each application service a generic test, which transparently sets up the SCS environment and feeds the service with test data.

%\textbf{Unexpected delays.} \comHE{TBD}

\textbf{Asset Administration Shells:} The exploitation of AAS in the context of lot size 1 production has proven very useful (\ref{qAAS}). AAS provide a manageable means to capture, update and check individualized production information across individual processes (in our case production and quality control) and across unit and organizational borders in an interoperable way\footnote{In the future, AAS formats that are currently in standardization will probably further facilitate interoperability)}. However, it has also shown that accompanying tools and infrastructure are crucial to make the use of AAS feasible (see also the discussion of BaSyx below). This is especially true, when AAS instances are required for individual products (not only for machines). In this case, an AAS editor, such as the frequently mentioned AASX package explorer\footnote{\url{https://github.com/admin-shell/aasx-package-explorer}}, may be helpful for AAS templates, but it is of little use for creating instances, as it is meant for supporting manual rather than for mass creation of AAS instances. Here, the automatic generation of AAS instances has proven to be a viable and effective way to create individual AAS based on production data. This is also very much in line with the low-code approach of our platform, which heavily relies on generated software artifacts (e.g., service code). 

\textbf{BaSyx Experience:} Overall, our experience with BaSyx was mostly positive, as it allows to effectively realize complex, distributed AAS-based applications or our platform AAS. However, it also became obvious that BaSyx is still in development and lacks some stability. For example, adding many elements (more than 1000 in a few seconds) to a submodel or removing such elements (e.g., by parallel processes) caused excessive CPU consumption. In our case, we mitigated these problems by aggressive filtering to focus on relevant information. Moreover, a more fine-grained modularization of BaSyx for application on resource-constrained devices would be desirable (although not causing issues on the same scale as, e.g., TensorFlow).

%We rely extensively on AAS, e.g., to continuously deliver traces of the internal processing to the tablet. Initial executions of the application surprisingly caused a CPU load of more than 400\% as well as server exceptions. An analysis showed that we accidentally pushed more than 1000 trace submodel collection entries within a few seconds into the AAS. Aggressive filtering to focus on relevant information as well as reduced connector sampling rates helped us mitigating this problem, while the exceptions remained. However, over time, approx. 40 trace entries per demonstrator run caused an excessive memory use that we could not attribute to other system components, e.g., to the robot control. As a consequence, we had to re-start the application container on the fair each 5-6 hours. Trials to automatically clean up the application AAS further affected system stability. Moreover, a more fine-grained modularization of BaSyx for application on resource-constrained devices would be desirable (although not causing issues on the same scale as, e.g., TensorFlow).  

\textbf{Platform Benefits:} Overall, using the IIP-Ecosphere platform allowed us to build the demonstrator in relatively short time (within four weeks from first conception, including planning, discussions, workarounds and bug fixing). To a large degree, we attribute this to the low-code model-based approach, which implies guidelines and structuring principles. For example, as an initial model of the application must exist before services can be implemented, the involved parties are motivated to agree on an application design first strengthening their joint vision and understanding (\ref{qInter}). From a technical and an organizational perspective, the service interfaces that we derived next, led to a clear separation of interests but also to ownerships, which contributes to \ref{qIntegrateTech} (here into AI, UI, robot control and supporting services). Moreover, the platform approach also gives structural guidance, e.g., on the organization of code artifacts for different programming languages so that packaged application artifacts can be deployed and executed by the platform. Further, the code generation helped us speeding up the development. In particular, the complete generation of OPC UA and AAS connectors from the model saved considerable amount of time. As already mentioned in Section~\ref{sect_useCase}, more than half of the application code was generated, relying on frameworks that were already tested and integrated into the platform. Further, the code generation even enabled us to quickly realize changes to the processing in an agile and consistent manner. Although the standardization of AAS is currently in progress, the existing AAS modeling concepts are already beneficial in data exchange as they lead to a uniform representation of data structures that are helpful, e.g., when realizing an application UI. A further benefit is that a platform provides an environment to realize services more easily, even without knowing details on machine connectors, data transport, monitoring or deployment. Here, an adequate abstraction of a functional unit, e.g., a service, allows to deploy and to distribute, e.g., AI, in a flexible and easy manner, e.g., to edge devices (\ref{qAI2edge}).

%\subsection{dataflow component testing}

%Camera tuning on the fly ;-)

As a more \textbf{general lesson} learned, it is definitively worth while to validate complex software systems such as a platform with realistic use cases, since it also brings improvements for the platform itself. Although our platform is subject to regression tests on different levels of granularity, the demonstrator involved several unexpected situations. For example, some of the control flow communication paths in the demonstrator that failed without (observable) reason. An initial workaround for the demonstrator finally led to a general revision of the code generation for these paths in the platform. Another example are services that need longer for their startup than expected by the tests, such as a Python script utilizing TensorFlow-light AI models. As the service startup was assigned to the wrong service lifecycle phase, these services were accidentally started multiple times by the service framework due to timeouts. Correcting the assignment for all services helped to further stabilize the platform.  

% ---------------------------------------------------------------------

%% file: main.bbl
\begin{thebibliography}{10}
\providecommand{\url}[1]{\texttt{#1}}
\providecommand{\urlprefix}{URL }
\providecommand{\doi}[1]{https://doi.org/#1}

\bibitem{angelopoulos2019tackling}
Angelopoulos, A., Michailidis, E.T., Nomikos, N., Trakadas, P., Hatziefremidis,
  A., Voliotis, S., Zahariadis, T.: Tackling faults in the industry 4.0 era—a
  survey of machine-learning solutions and key aspects. Sensors
  \textbf{20}(1), ~109 (2019)

\bibitem{genFrame11}
Bader, S., Bedenbecker, H., Billmann, M., et~al.: {Generic Frame for Technical
  Data for Industrial Equipment in Manufacturing (Version 1.1)} (2020),
  \url{https://bit.ly/3NpuIOF}

\bibitem{bansod2021analysis}
Bansod, G., Khandekar, S., Khurana, S.: Analysis of convolution neural network
  architectures and their applications in industry 4.0. In: Metaheuristic
  Algorithms in Industry 4.0, pp. 139--162. CRC Press (2021)

\bibitem{aasMonitoring21}
Casado, M.G., Eichelberger, H.: {Industry 4.0 Resource Monitoring - Experiences
  With Micrometer and Asset Administration Shells}. In: CEUR-WS Proceedings of
  Symposium on Software Performance 2021 (SSP'21) (2021)

\bibitem{fedmodels}
Chang, Y., Laridi, S., Ren, Z., Palmer, G., Schuller, B.W., Fisichella, M.:
  Robust federated learning against adversarial attacks for speech emotion
  recognition. arXiv preprint arXiv:2203.04696  (2022)

\bibitem{chen2021artificial}
Chen, J., Lim, C.P., Tan, K.H., Govindan, K., Kumar, A.: Artificial
  intelligence-based human-centric decision support framework: an application
  to predictive maintenance in asset management under pandemic environments.
  Annals of Operations Research pp. 1--24 (2021)

\bibitem{chen_iot_2020}
Chen, S., Li, Q., Zhang, H., Zhu, F., Xiong, G., Tang, Y.: An {IoT} {Edge}
  {Computing} {System} {Architecture} and its {Application}. In:
  {International} {Conference} on {Networking}, {Sensing} and {Control}
  ({ICNSC}). pp.~1--7 (2020)

\bibitem{denkena2021scalable}
Denkena, B., Dittrich, M.A., Fohlmeister, S., Kemp, D., Palmer, G.: Scalable
  cooperative multi-agent-reinforcement-learning for order-controlled on
  schedule manufacturing in flexible manufacturing systems. Simulation in
  Produktion und Logistik 2021: Erlangen, 15.-17. September 2021 p.~305 (2021)

\bibitem{EichelbergerQinSizonenko+16}
Eichelberger, H., Qin, C., Sizonenko, R., Schmid, K.: {Using IVML to Model the
  Topology of Big Data Processing Pipelines}. In: International Systems and
  Software Product Line Conference. pp. 204 -- 208 (2016)

\bibitem{ESS22}
Eichelberger, H., Stichweh, H., Sauer, C.: {Requirements for an AI-enabled
  Industry 4.0 Platform – Integrating Industrial and Scientific Views}. In:
  Intl. Conference on Advances and Trends in Software Engineering. pp. 7--14
  (2022)

\bibitem{FeichtingerMeixnerRinker+22}
Feichtinger, K., Meixner, K., Rinker, F., Koren, I., et~al.: {Industry Voices
  on Software Engineering Challenges in Cyber-Physical Production Systems
  Engineering}. In: ETFA'22 (2022), \url{https://t.co/sqr3XfVMyN}, accepted,
  preprint

\bibitem{foukalas_cognitive_2020}
Foukalas, F.: Cognitive {IoT} platform for fog computing industrial
  applications. Computers \& Electrical Engineering  \textbf{87},  106770
  (2020)

\bibitem{han2021pre}
Han, X., Zhang, Z., Ding, N., Gu, Y., Liu, X., Huo, Y., Qiu, J., Yao, Y.,
  Zhang, A., Zhang, L., et~al.: Pre-trained models: Past, present and future.
  AI Open  \textbf{2},  225--250 (2021)

\bibitem{HOANG201942}
Hoang, D.T., Kang, H.J.: Rolling element bearing fault diagnosis using
  convolutional neural network and vibration image. Cognitive Systems Research
  \textbf{53},  42--50 (2019), advanced Intelligent Computing

\bibitem{howard2017mobilenets}
Howard, A.G., Zhu, M., Chen, B., Kalenichenko, D., Wang, W., Weyand, T.,
  Andreetto, M., Adam, H.: {Mobilenets: Efficient convolutional neural networks
  for mobile vision applications}. arXiv preprint arXiv:1704.04861  (2017)

\bibitem{HEG+18}
Hummel, O., Eichelberger, H., Giloj, A., Werle, D., Schmid, K.: {A Collection
  of Software Engineering Challenges for Big Data System Development}. In:
  Euromicro Conf. on Software Engineering and Advanced Applications. pp.
  362--369 (2018)

\bibitem{KannothHermannDamm+21}
Kannoth, S., Hermann, J., Damm, M., R{\"u}bel, P., Rusin, D., Jacobi, M.,
  Mittelsdorf, B., Kuhn, T., Antonino, P.O.: {Enabling SMEs to Industry 4.0
  Using the BaSyx Middleware: A Case Study}. In: Software Architecture. pp.
  277--294 (2021)

\bibitem{kiangala2020effective}
Kiangala, K.S., Wang, Z.: An effective predictive maintenance framework for
  conveyor motors using dual time-series imaging and convolutional neural
  network in an industry 4.0 environment. Ieee Access  \textbf{8},
  121033--121049 (2020)

\bibitem{platfIAO}
Krause, T., Strauß, O., Scheffler, G., Kett, H., Lehmann, K., Renner, T.:
  {IT-Plattformen für das Internet der Dinge (IoT)} (2017),
  \url{https://bit.ly/3amlXaG}

\bibitem{lecun2015deep}
LeCun, Y., Bengio, Y., Hinton, G.: Deep learning. nature  \textbf{521}(7553),
  436--444 (2015)

\bibitem{lee2019quality}
Lee, S.M., Lee, D., Kim, Y.S.: The quality management ecosystem for predictive
  maintenance in the industry 4.0 era. International Journal of Quality
  Innovation  \textbf{5}(1),  1--11 (2019)

\bibitem{lin2022human}
Lin, C.H., Wang, K.J., Tadesse, A.A., Woldegiorgis, B.H.: Human-robot
  collaboration empowered by hidden semi-markov model for operator behaviour
  prediction in a smart assembly system. Journal of Manufacturing Systems
  \textbf{62},  317--333 (2022)

\bibitem{LSR07}
van~der Linden, F., Schmid, K., Rommes, E.: {Software Product Lines in Action -
  The Best Industrial Practice in Product Line Engineering}. Springer (2007)

\bibitem{lins_industry_2018}
Lins, T., Oliveira, R.A.R., Correia, L.H.A., Silva, J.S.: {Industry 4.0
  {Retrofitting}}. In: {Brazilian} {Symp.} on {Computing} {Systems}
  {Engineering} ({SBESC}). pp. 8--15 (2018)

\bibitem{iipSurvey}
Niederée, C., Eichelberger, H., Schmees, H.D., Broos, A., Schreiber, P.: {KI
  in der Produktion – Quo vadis?} (2021),
  \url{https://doi.org/10.5281/zenodo.6334521}

\bibitem{pal2021deep}
Pal, S.K., Pramanik, A., Maiti, J., Mitra, P.: Deep learning in multi-object
  detection and tracking: state of the art. Applied Intelligence
  \textbf{51}(9),  6400--6429 (2021)

\bibitem{palmer2020automated}
Palmer, G., Schnieders, B., Savani, R., Tuyls, K., Fossel, J., Flore, H.: The
  automated inspection of opaque liquid vaccines. In: ECAI 2020, pp. 1898--1905
  (2020)

\bibitem{patalas2020ai}
Patalas-Maliszewska, J., Paj{k{a}}k, I., Skrzeszewska, M.: Ai-based
  decision-making model for the development of a manufacturing company in the
  context of industry 4.0. In: International Conference on Fuzzy Systems
  (FUZZ-IEEE). pp.~1--7 (2020)

\bibitem{peres2020industrial}
Peres, R.S., Jia, X., Lee, J., Sun, K., Colombo, A.W., Barata, J.: Industrial
  artificial intelligence in industry 4.0 - systematic review, challenges and
  outlook. IEEE Access  \textbf{8},  220121--220139 (2020)

\bibitem{DetailsAAS}
{{Platform Industrie 4.0}}: {Details of the Asset Administration Shell},
  \url{https://bit.ly/3nPPAUU}

\bibitem{raileanu_edge_2018}
Raileanu, S., Borangiu, T., Morariu, O., Iacob, I.: Edge {Computing} in
  {Industrial} {IoT} {Framework} for {Cloud}-based {Manufacturing} {Control}.
  In: {International} {Conference} on {System} {Theory}, {Control} and
  {Computing} ({ICSTCC}). pp. 261--266 (2018)

\bibitem{reimer2022identifying}
Reimer, J., Wang, Y., Laridi, S., Urdich, J., Wilmsmeier, S., Palmer, G.:
  Identifying cause-and-effect relationships of manufacturing errors using
  sequence-to-sequence learning. arXiv preprint arXiv:2205.02827  (2022)

\bibitem{iipPlatfOver}
Sauer, C., Eichelberger, H., Ahmadian, A.S., Dewes, A., Jürjens, J.: {Current
  Industrie 4.0 Platforms – An Overview} (2021),
  \url{https://zenodo.org/record/4485756}

\bibitem{schnieders2019fully}
Schnieders, B., Luo, S., Palmer, G., Tuyls, K.: Fully convolutional one-shot
  object segmentation for industrial robotics. In: International Conference on
  Autonomous Agents and MultiAgent Systems. pp. 1161--1169 (2019)

\bibitem{SculleyHoltGolovin15}
Sculley, D., Holt, G., Golovin, D., Davydov, E., Phillips, T., Ebner, D.,
  Chaudhary, V., Young, M., Crespo, J.F., Dennison, D.: {Hidden Technical Debt
  in Machine Learning Systems}. In: Advances in Neural Information Processing
  Systems. vol.~28, pp. 2503--2511 (2015)

\bibitem{shinde2018review}
Shinde, P.P., Shah, S.: A review of machine learning and deep learning
  applications. In: International conference on computing communication control
  and automation (ICCUBEA). pp.~1--6 (2018)

\bibitem{silva2019assets}
Silva, W., Capretz, M.: Assets predictive maintenance using convolutional
  neural networks. In: International Conference on Software Engineering,
  Artificial Intelligence, Networking and Parallel/Distributed Computing
  (SNPD). pp. 59--66 (2019)

\bibitem{WANG2019182}
Wang, H., Li, S., Song, L., Cui, L.: A novel convolutional neural network based
  fault recognition method via image fusion of multi-vibration-signals.
  Computers in Industry  \textbf{105},  182--190 (2019)

\bibitem{zhuang2018digital}
Zhuang, C., Liu, J., Xiong, H.: Digital twin-based smart production management
  and control framework for the complex product assembly shop-floor. The
  international journal of advanced manufacturing technology  \textbf{96}(1),
  1149--1163 (2018)

\end{thebibliography}
